\documentclass[pdflatex,sn-mathphys-num]{sn-jnl}% Math and Physical Sciences Numbered Reference Style
\usepackage{graphicx}
\usepackage{multirow}
\usepackage{amsmath,amssymb,amsfonts}
\usepackage{amsthm}
\usepackage{mathrsfs}
\usepackage[title]{appendix}
\usepackage{xcolor}
\usepackage{textcomp}
\usepackage{manyfoot}
\usepackage{booktabs}
\usepackage{algorithm}
\usepackage{algorithmicx}
\usepackage{algpseudocode}
\usepackage{listings}
\usepackage{orcidlink}
\usepackage{array}
\usepackage{colortbl}
\usepackage{stfloats}
\usepackage{url}
\usepackage{verbatim}
\usepackage{tikz}
\usepackage{balance}
\usepackage{graphics}
\usepackage{epsfig}
\usepackage{times}
\usepackage{siunitx}
\usepackage{hyperref}
\usepackage{multicol}
\usepackage{nicefrac}
\usepackage{svg}
\usepackage{picinpar}
\usepackage[latin1]{inputenc}
\usepackage{soul}
\usepackage{pifont}
\usepackage{color}
\usepackage{alltt}
\usepackage{enumerate}
\usepackage{epstopdf}
\usepackage{pbox}
\usepackage{diagbox}
\usepackage{microtype}
\usepackage{subfigure}
\usepackage{booktabs}
\usepackage{xfrac}
\usepackage{gensymb}
\usepackage{xr}
\usepackage{makecell}

\theoremstyle{thmstyleone}%
\theoremstyle{thmstyletwo}%

\theoremstyle{thmstylethree}%

\DeclareMathOperator*{\argmin}{arg\,min}

\begin{document}

\title[Article Title]{Learning High-Quality Latent Representations for Anomaly Detection and Signal Integrity Enhancement in High-Speed Signals}

\author[1]{\fnm{Muhammad} \sur{Usama}\orcidlink{0000-0002-3834-2167}}\email{usama@kaist.ac.kr}

\author[1]{\fnm{Hee-Deok} \sur{Jang}\orcidlink{0000-0001-5153-908X}}\email{jhd6844@kaist.ac.kr}

\author[1]{\fnm{Soham} \sur{Shanbhag}\orcidlink{0000-0002-5541-2016}}\email{sshanbhag@kaist.ac.kr}

\author[2]{\fnm{Yoo-Chang} \sur{Sung}}\email{yc.sung@samsung.com}

\author[2]{\fnm{Seung-Jun} \sur{Bae}\orcidlink{0000-0003-0077-7488}}\email{seungjun.bae@samsung.com}

\author*[1]{\fnm{Dong Eui} \sur{Chang}\orcidlink{0000-0002-6496-4189}}\email{dechang@kaist.ac.kr}

\affil*[1]{\orgdiv{School of Electrical Engineering}, \orgname{KAIST}, \orgaddress{\city{Daejeon}, \postcode{34141}, \country{Republic of Korea}}}

\affil[2]{\orgdiv{DRAM Design Team, Memory Division}, \orgname{Samsung Electronics}, \orgaddress{\city{Hwasung}, \country{Republic of Korea}}}

% \funding{This work was supported by Samsung Electronics Co. Ltd (Contract ID : MEM230315\_0004).}


\abstract{This paper addresses the dual challenge of improving anomaly detection and signal integrity in high-speed dynamic random access memory signals. To achieve this, we propose a joint training framework that integrates an autoencoder with a classifier to learn more distinctive latent representations by focusing on valid data features. 
Our approach is evaluated across three anomaly detection algorithms and consistently outperforms two baseline methods. Detailed ablation studies further support these findings. Furthermore, we introduce a signal integrity enhancement algorithm that improves signal integrity by an average of 11.3\%. The source code and data used in this study are available at \url{https://github.com/Usama1002/learning-latent-representations}.}

\keywords{Anomaly Detection, Signal Integrity, Representation Learning, Machine Learning, Autoencoders}



\maketitle

\section{Introduction}

Dynamic random access memory (DRAM) systems are critical for modern computing infrastructure, requiring robust signal transmission to prevent data corruption and system failures. Anomaly detection and signal integrity (SI) monitoring are essential for DRAM reliability, especially in high-performance computing where minor signal degradations can cause significant system-wide issues.

Detecting anomalies in DRAM signals is challenging due to subtle signal degradations and high-dimensional, time-series characteristics. Anomalies manifest as voltage deviations, timing variations, or noise patterns that are difficult to distinguish from normal variations. The dynamic nature of DRAM operations means normal patterns vary significantly with operating conditions, memory access patterns, and environmental factors, making fixed thresholds or rule-based systems unreliable.

DRAM anomaly detection has evolved from traditional rule-based and threshold-based methods \cite{liu2013experimental, kim2014flipping, pattabiraman2008symplfied} to machine learning-based solutions. Recent research uses trained encoders and feature extraction methods for anomaly detection algorithms \cite{ml_journal_paper, ml_paper, valid_data_autoencoder, Gao2023TSMAEAN}. Autoencoder-based approaches learn latent representations of error-free circuit simulations for detecting anomalies in waveform data \cite{ml_journal_paper, ml_paper}, while memory-augmented autoencoders capture temporal dependencies \cite{Gao2023TSMAEAN, memAE}.

Current methods have significant limitations. Trained encoders excel at general feature extraction but struggle with task-specific features crucial for DRAM anomaly detection \cite{palakurti2024challenges, ZhouPaffenroth, Malhotra2015LongST, chen}. Unsupervised autoencoders that train on normal data and use reconstruction error \cite{valid_data_autoencoder} lack discriminative power to distinguish subtle normal variations from genuine anomalies.

Semi-supervised approaches combining autoencoders with supervised classifiers have been proposed \cite{ZhouPaffenroth, chen, akcay2019ganomaly}, using classifier loss to guide learning of discriminative representations. However, obtaining labeled anomalous DRAM data is extremely difficult, as anomalies are rare and hard to identify. Our experiments and previous work \cite{chen, SemiSupervisedLB, memAE} show that using labeled anomalous data causes overfitting and reduced performance on diverse anomaly types.

We propose a semi-supervised approach that trains an autoencoder and classifier simultaneously while backpropagating gradients only from normal data through the classifier. This addresses overfitting while learning discriminative latent representations that capture normal data patterns. By jointly optimizing autoencoder reconstruction loss and classifier objective on normal data only, our method learns representations that are both reconstructive and discriminative.

Our approach significantly improves over existing methods by combining autoencoder reconstruction with classifier discrimination while avoiding pitfalls of training on limited anomalous samples. The learned representations demonstrate superior performance detecting various DRAM signal anomalies compared to traditional autoencoder-based and other machine learning methods.

We also propose a signal integrity enhancement algorithm that leverages our learned latent representations to improve DRAM signal quality. This SI enhancement capability is unique, as existing anomaly detection methods focus solely on detection without remediation. Our approach identifies anomalies and provides signal improvement, validated through eye diagram analysis demonstrating measurable SI enhancements. This enhancement algorithm also shows the improvements afforded by our proposed training method to learn high-quality latent representations.

% An additional benefit is our signal integrity enhancement algorithm that leverages learned latent representations to improve DRAM signal quality. This SI enhancement capability is unique, as existing anomaly detection methods focus solely on detection without remediation. Our approach identifies anomalies and provides signal improvement, validated through eye diagram analysis demonstrating measurable SI enhancements.
\section{Details about the Dataset}
\label{sec:dataset}
\begin{figure}[t]
    \centering
    \includegraphics[width=.8\textwidth]{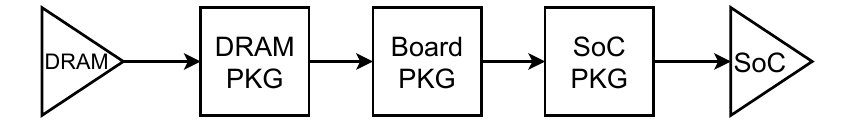}
    \caption{The structure of the channel for waveform generation.}
    \label{fig:dram_pipeline}
\end{figure}

In this section, we provide an overview of the data utilized in our study, including its collection process and subsequent preprocessing steps. We collect experimental data to evaluate the signal integrity using simulation modeling of a mobile DRAM channel. The channel structure modeled in the simulation is shown in Figure~\ref{fig:dram_pipeline}. DRAM and system-on-chip (SoC), represented as triangular shapes in Figure~\ref{fig:dram_pipeline}, act as data transmitter and receiver, respectively. DRAM generates a pseudorandom binary sequence pattern and transmits it to SoC at a speed of \SI{10}{Gbps}. We set the rise time for waveform generation at \SI{30}{\pico\second} and record the received waveforms with a minimum sampling time of \SI{1}{\pico\second}. Furthermore, to observe different signal characteristics, we generate waveforms in five different cases, which are given in Table~\ref{tab:test-cases}.

\begin{table}[t]
    \caption{Data Collection Test Cases Configurations}
    \label{tab:test-cases}
    \centering
    \begin{tabular}{ccccc}
    \toprule
    \multirow{2}{*}{Case} & Channel & Driver & On-Die & Pattern \\
    & Configuration & Strength & Termination & Type \\
    \midrule
    1 & DRAM package only & \SI{80}{\ohm} & \SI{80}{\ohm} & Single \\
    % \hline
    2 & DRAM + Board + SoC & \SI{80}{\ohm} & \SI{80}{\ohm} & Single \\
    % \hline
    3 & DRAM + Board + SoC & \SI{240}{\ohm} & $\infty$ & Single \\
    % \hline
    4 & DRAM + Board + SoC & \SI{240}{\ohm} & \SI{240}{\ohm} & Single \\
    % \hline
    5 & DRAM + Board + SoC & \SI{240}{\ohm} & \SI{240}{\ohm} & Multiple \\
    \bottomrule
    \end{tabular}
\end{table}

The collected experimental data undergoes a series of preprocessing steps. This involves selecting the waveform data while excluding the initial transient portion, $\approx$\SI{3000}{\pico\second} in duration. Although the recorded waveform data have a minimum sampling time of \SI{1}{\pico\second}, there are variations in the time intervals between the recorded values. To standardize the dataset, time intervals between recorded values are adjusted using linear interpolation to \SI{1}{\pico\second}.

\begin{figure*}[t]
	\centering
	\subfigure[Example of an invalid signal ($y=0$)]
	{
		\includegraphics[width=.45\textwidth]{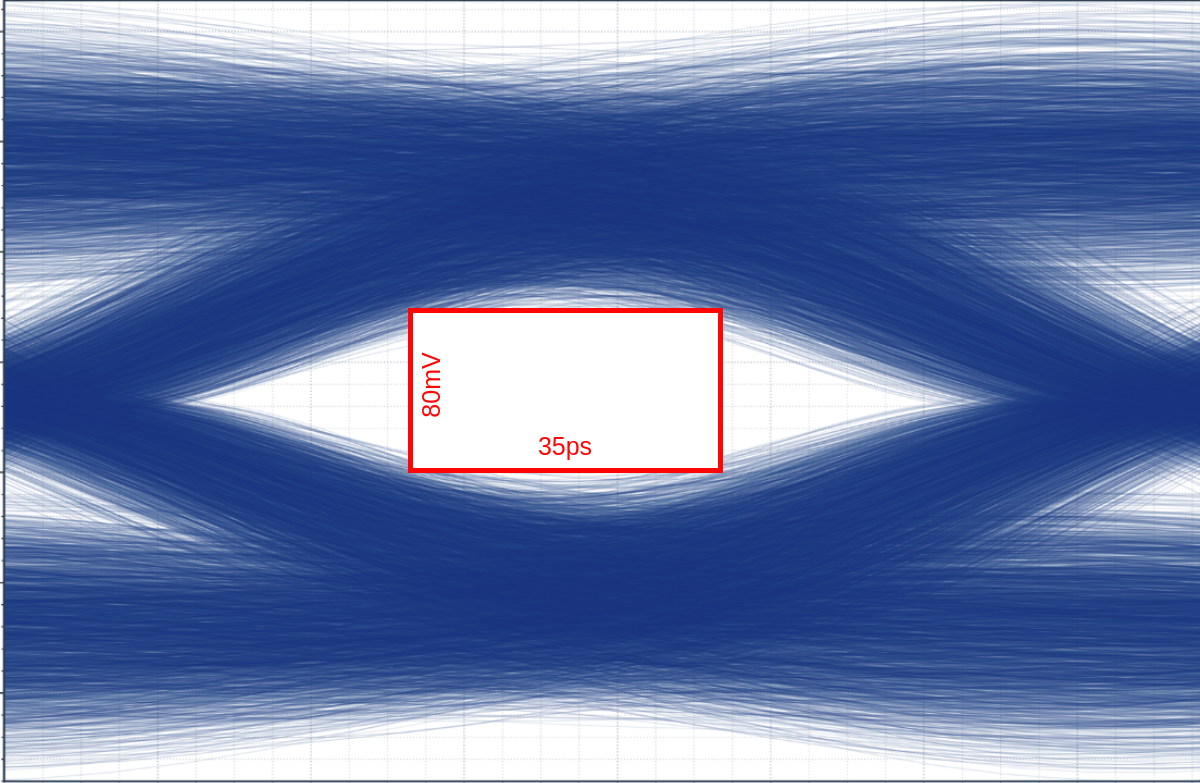}
	}
	\subfigure[Example of a valid signal ($y=1$)]
	{
		\includegraphics[width=.45\textwidth]{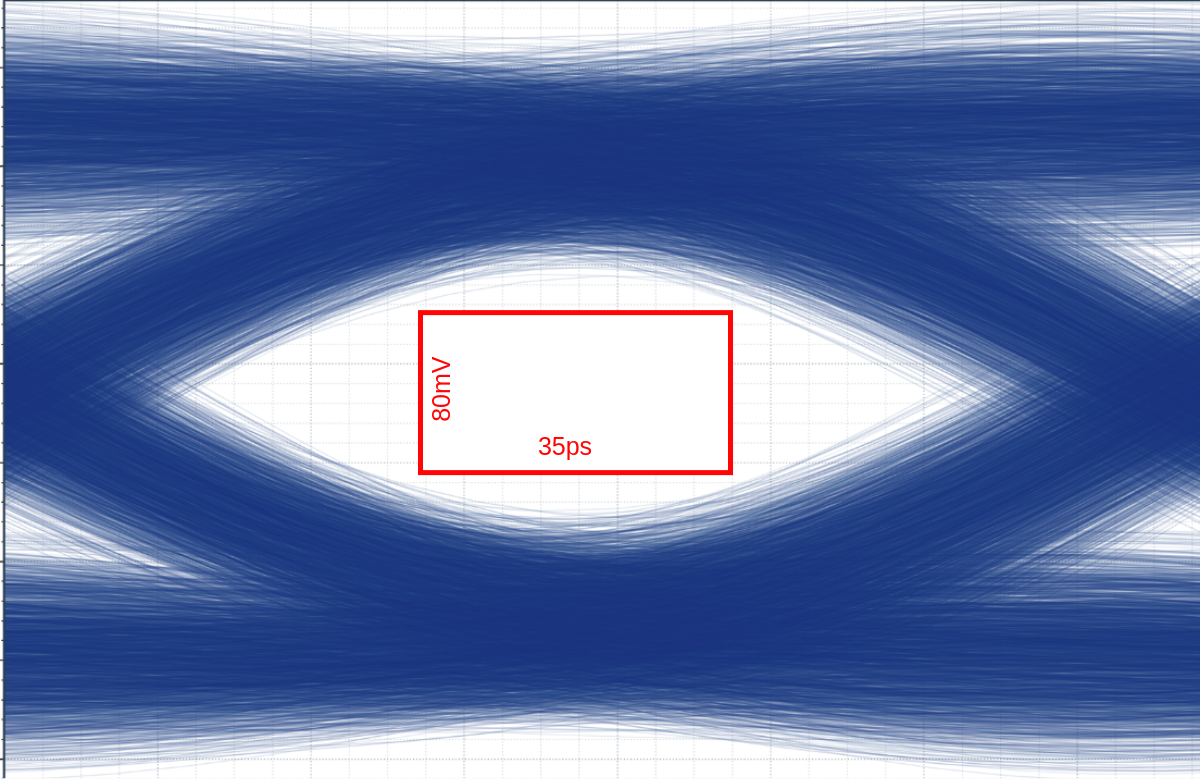}
	}
	\caption{Illustration of the signal validity labeling criteria. The rectangular window (80 mV $\times$ 35 ps) is shown in red. (a) An invalid signal where signal transitions intersect the window, and (b) a valid signal where no transitions occur within the window region.}
	\label{fig:data_labeling}
\end{figure*}

We use eye diagram analysis to classify signals as valid or invalid. Eye diagrams are constructed by superimposing rising and falling edges across 100 unit intervals of \SI{100}{\pico\second} each, yielding 10,000 data points total. A rectangular window measuring 80 mV in height and 35 ps in width is fit within the eye-opening \cite{PyEye}. If the signal intersects this window, it is labeled as invalid (\(y=0\)); otherwise, it is considered valid (\(y=1\)), as illustrated in Figure~\ref{fig:data_labeling}. Each signal is divided into $n_x = 100$ length vectors, such that the training data consists of vectors $\mathbf{x} \in \mathbb{R}^{n_x}$, with all vectors from the same signal sharing the same label $y \in \{0,1\}$ (valid: $y=1$, invalid: $y=0$). In our case, each signal is divided into 100 vectors.

To address class imbalance favoring valid data, we generate synthetic invalid samples by applying distortions (inter-signal interference, amplitude distortion, harmonic distortion) to valid samples, mimicking real-world semiconductor signal anomalies. The combined dataset is randomly split into \SI{80}{\percent} training and \SI{20}{\percent} testing.

\section{Main Contributions}
\label{sec:method}
This section details our approach to DRAM anomaly detection and signal integrity (SI) enhancement. We present: (1) a novel training method for learning discriminative latent representations for anomaly detection, and (2) an algorithm for SI enhancement.

\subsection{Preliminaries and Notation}
Let $\mathbf{x} \in \mathbb{R}^{n_x}$ denote an input signal vector of dimension $n_x$, and $y \in \{0,1\}$ denote its corresponding binary label, where $y=1$ indicates a valid signal and $y=0$ indicates an anomalous signal. We define three neural networks: an encoder $\mathbf{E}: \mathbb{R}^{n_x} \rightarrow \mathbb{R}^l$ that maps input signals to latent vectors of dimension $l$, a decoder $\mathbf{D}: \mathbb{R}^l \rightarrow \mathbb{R}^{n_x}$ that reconstructs signals from their latent representations, and a classifier $\mathbf{C}: \mathbb{R}^l \rightarrow [0,1]$ that predicts the probability of a signal being valid based on its latent representation. For any signal $\mathbf{x}$, we denote its latent representation as $\mathbf{z} = \mathbf{E}(\mathbf{x})$, its reconstruction as $\hat{\mathbf{x}} = \mathbf{D}(\mathbf{z})$, and its predicted label (valid/invalid) as $\hat{y} = \mathbf{C}(\mathbf{z})$. Throughout, $\|\cdot\|$ denotes the Euclidean $(L_2)$ norm.

Let $\mathcal{X}_v$ and $\mathcal{X}_a$ denote the set of all valid and anomalous signals, respectively. We define $\mathbf{c}\in\mathbb{R}^{l}$ as the anchor point of the valid data latent representations and is obtained by solving the following Fermat-Weber optimization problem over $\{\mathbf{E}(\mathbf{x}): \mathbf{x}\in\mathcal{X}_v\}$, i.e.,
\begin{equation}
    \mathbf{c}=\argmin_{\mathbf{k}\in\mathbb{R}^{l}} \sum_{\mathbf{x}\in\mathcal{X}_v}\|\mathbf{E}(\mathbf{x})-\mathbf{k}\|.
    \label{eq:fermat}
\end{equation}

\begin{figure}[t]
    \centering
    \includesvg[width=1.1\linewidth]{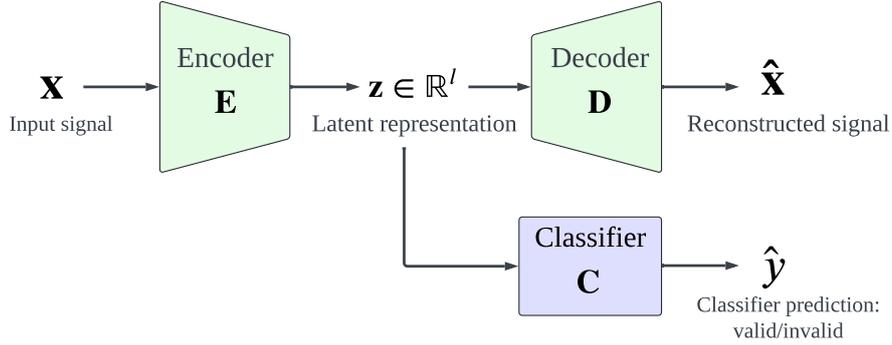}
    \caption{Structure of the proposed training method, which includes an autoencoder model combined with a classification network.}
    \label{fig:concept}
\end{figure}

\subsection{Latent Representation Learning for Enhanced Anomaly Detection}
\label{sec:training_method}
We propose a method combining an autoencoder and a classifier to extract latent vectors from high-speed DRAM signals for anomaly detection. While prior work uses autoencoder encoders as feature extractors \cite{ml_journal_paper, ml_paper, valid_data_autoencoder}, their reconstruction focus can overlook critical anomaly-specific features \cite{palakurti2024challenges, professor_book, representation_learning_review, hinton2006reducing}. Supervised methods struggle with obtaining comprehensive anomalous data, risking overfitting and poor generalization \cite{chen, SemiSupervisedLB, memAE}. Similarly, semi-supervised methods using limited anomalous data for classifiers can overfit. Our approach integrates a classifier with an autoencoder, training primarily on normal data to learn robust representations and mitigate these issues.

Figure~\ref{fig:concept} visualizes our structure. The autoencoder network maps the input $\mathbf{x}$ to an $l=11$ dimensional latent space, and a classifier operates on these latent vectors to predict the signal validity. Table~\ref{tab:network-arch} shows the network architectures. Both networks are trained simultaneously by minimizing the loss function
\begin{equation}
\label{eq:loss}
    \mathcal{L} = \Vert \mathbf{x} - \hat{\mathbf{x}} \Vert^{2} - y\log(\hat{y}).
\end{equation}
Crucially, only valid data gradients (i.e., with $y=1$) are backpropagated from the classifier network. This encourages tight clustering of valid signal representations and separation of anomalies. 

Unlike \cite{b2}'s approach which uses a linear classifier and both valid and anomalous gradients, our method exclusively uses valid data gradients. Our method addresses the impracticality of collecting comprehensive anomaly data and leads to better generalization by mitigating overfitting to specific anomaly types \cite{chen, SemiSupervisedLB, memAE}. Focusing on valid data patterns enhances identification of diverse deviations. The trained encoder serves as our feature extractor. Its latent vectors are used by three anomaly detection algorithms: unsupervised LOF \cite{lof}, semi-supervised LSAnomaly \cite{lsanomaly}, and a fully supervised neural network (three fully-connected layers with ReLU activation in the first two layers, and sigmoid activation in the output).

% \begin{table}[t]
%     \caption{Network Architecture Details}
%     \label{tab:network-arch}
%     \centering
%     \begin{tabular}{|l|l|c|c|c|}
%     \hline
%     \multirow{2}{*}{Network} & \multirow{2}{*}{Layer} & \multicolumn{3}{c|}{Architecture Details} \\
%     \cmidrule{3-5}
%     & & Input Size & Output Size & Activation \\
%     \hline
%     \multirow{5}{*}{Encoder} & Input & $n_x=100$ & 512 & ReLU \\
%     & FC & 512 & 256 & ReLU \\
%     & FC & 256 & 128 & ReLU \\
%     & FC & 128 & 64 & ReLU \\
%     & FC & 64 & $l=11$ & ReLU \\
%     \hline
%     \multirow{5}{*}{Decoder} & FC & $l=11$ & 64 & ReLU \\
%     & FC & 64 & 128 & ReLU \\
%     & FC & 128 & 256 & ReLU \\
%     & FC & 256 & 512 & ReLU \\
%     & Output & 512 & $n_x=100$ & 1.5*Tanh \\
%     \hline
%     Classifier & FC & $l=11$ & 1 & Sigmoid \\
%     \hline
%     \end{tabular}
% \end{table}

\begin{table}[t]
\caption{Network Architecture Details}
\label{tab:network-arch}
\centering
\begin{tabular}{@{}llccc@{}}
\toprule
Network & Layer & Input Size & Output Size & Activation \\
\midrule
\multirow{5}{*}{Encoder} 
    & Input & $n_x=100$ & 512 & ReLU \\
    & FC & 512 & 256 & ReLU \\
    & FC & 256 & 128 & ReLU \\
    & FC & 128 & 64 & ReLU \\
    & FC & 64 & $l=11$ & ReLU \\
\midrule
\multirow{5}{*}{Decoder} 
    & FC & $l=11$ & 64 & ReLU \\
    & FC & 64 & 128 & ReLU \\
    & FC & 128 & 256 & ReLU \\
    & FC & 256 & 512 & ReLU \\
    & Output & 512 & $n_x=100$ & 1.5*Tanh \\
\midrule
Classifier & FC & $l=11$ & 1 & Sigmoid \\
\botrule
\end{tabular}
\end{table}

\subsection{SI Enhancement Algorithm Using Latent Representations}
\label{sec:enhancing_signal_integrity_si_metric}
Leveraging the learned representations, we propose an SI enhancement algorithm. We define an SI metric $\sigma(\mathbf{x}) = \|\mathbf{E}(\mathbf{x}) - \mathbf{c}\|$, where $\mathbf{c}$ is the Fermat-Weber point (Eq. \eqref{eq:fermat}), to quantify signal integrity in the latent space.

Our algorithm assumes lower $\sigma(\mathbf{x})$ indicates higher SI, reflecting better alignment with valid data characteristics. For a given signal $\mathbf{x}$, the algorithm proceeds as follows:

\begin{enumerate}[Step 1:]
    \item Compute latent vector $\mathbf{z} = \mathbf{E}(\mathbf{x})$; initialize enhanced signal $\mathbf{x}_e = \mathbf{x}$.
    \item Iteratively move $\mathbf{z}$ towards $\mathbf{c}$ in $m$ steps: $\mathbf{z}_t = (1 - \frac{t}{m})\mathbf{z} + \frac{t}{m}\mathbf{c}$ for $t \in [0,m]$.
    \item At each step $t$, reconstruct signal $\hat{\mathbf{x}}_t = \mathbf{D}(\mathbf{z}_t)$. If its dissimilarity $d(\mathbf{x}, \hat{\mathbf{x}}_t)$ from the original $\mathbf{x}$ given by
    \begin{equation}
    \label{eq:dissimilarity}
            d(\mathbf{x}, \hat{\mathbf{x}}_t) = \sqrt{\frac{1}{n_x} \sum_{i = 1}^{n_x} (x_i - \hat{x}_{t,i})^2},
    \end{equation}    
    is less than $\alpha$ (hyperparameter), update $\mathbf{x}_e \leftarrow \hat{\mathbf{x}}_t$.
\end{enumerate}
The pseudocode for our algorithm is given in Algorithm~\ref{alg:algorithm}.

\begin{algorithm}[t]
\caption{Algorithm to enhance the signal integrity.}
\label{alg:algorithm}
\begin{algorithmic}
\State \textbf{Input:} Given signal $\mathbf{x}$, anchor $\mathbf{c}$, trained networks $\mathbf{E}$ and $\mathbf{D}$ 
\State Set hyperparameter $\alpha \in \mathbb{R}_{\geq 0}$ as the desired threshold
\State Obtain latent vector: $\mathbf{z} \gets \mathbf{E}(\mathbf{x})$
\State Initialize step size: $m \gets \lceil \frac{\|\mathbf{z}-\mathbf{c}\|}{\alpha/L_D} \rceil$
\State Initialize enhanced signal: $\mathbf{x}_e \gets \mathbf{x}$
\State
\For{$t = 0$ \textbf{to} $m$}
    \State Update latent vector: $\mathbf{z}_t \gets (1 - \frac{t}{m})\mathbf{z} + \frac{t}{m}\mathbf{c}$
    \State Reconstruct signal: $\hat{\mathbf{x}}_t \gets \mathbf{D}(\mathbf{z}_t)$
    \State Calculate dissimilarity $d(\mathbf{x}, \hat{\mathbf{x}}_t)$ as given in equation \eqref{eq:dissimilarity}
    \If{$d(\mathbf{x}, \hat{\mathbf{x}}_t) < \alpha$}
        \State $\mathbf{x}_e \gets \hat{\mathbf{x}}_t$
    \Else
        \State \textbf{break}
    \EndIf
\EndFor
\State
\State \textbf{Output:} Enhanced signal: $\mathbf{x}_e$ (with improved signal integrity)
\end{algorithmic}
\end{algorithm}
\section{Experimental Setup and Results}
\label{sec:experiments}

\subsection{Training Details and Experimental Settings}
\label{sec:model_details}
We use the Adam optimizer to simultaneously train the autoencoder and classifier networks, with Xavier initialization for network parameters. The initial learning rate is set to 0.05 and gradually decreased using an exponential schedule with a decay rate of 0.75. Both the autoencoder and classifier are trained simultaneously for 100 epochs and a batch size of 100. 

We evaluate the performance of latent vectors generated by the trained encoder obtained by the proposed training method in comparison to two baseline approaches. Baseline 1 \cite{ml_journal_paper} uses the contractive autoencoder \cite{contractiveAE} and the loss function
\begin{equation*}
        \mathcal{L}_{\text{Baseline 1}} = \Vert \mathbf{x} - \hat{\mathbf{x}} \Vert^{2} + \lambda \| \mathbf{J}_e(\mathbf{x}) \|^2_{F},
\end{equation*}
where $\lambda \in \mathbb{R}_{> 0}$ is a constant and $\| \mathbf{J}_e(\mathbf{x}) \|^2_{F} = \sum_{i,j} (\nicefrac{\partial z_j(\mathbf{x})}{\partial x_i})^2$, where $z_j(\mathbf{x})$ gives the $j$-th element in the latent vector $\mathbf{z}$ and $x_i$ gives the $i$-th element in the input vector $\mathbf{x}$. In our experiments, we use $\lambda = 10^{-4}$.
Baseline 2 is identical to our proposed method except it uses gradients of both valid and invalid data samples during training. 
The loss function for training Baseline 2 is given as
\begin{equation}
    \mathcal{L}_{\text{Baseline 2}} = \Vert \mathbf{x} - \hat{\mathbf{x}} \Vert^{2} - y \log(\hat{y}) - (1 - y) \log(1 - \hat{y}). \nonumber
\end{equation}
All the methods share identical training conditions and autoencoder architecture.

\subsection{Enhanced Anomaly Detection Performance}
\label{sec:exp:classification_performance}
We employ three algorithms for anomaly detection, LoF, LSAnomaly and neural network, described in Section \ref{sec:training_method}. 
All three anomaly detectors are trained using the latent representations generated by the frozen encoder. Table \ref{tab:anomaly_detection_results} compares the anomaly detection performance of the latent representations learned using our proposed method with those learned using baseline training methods. 
Our method consistently outperforms the baseline methods across all algorithms. Notably, the superior performance of our method over Baseline 1, which does not utilize a classifier network, highlights the importance of incorporating the classifier network for obtaining distinctive latent representations for anomaly detection. Additionally, the performance improvement of our method over Baseline 2 shows the detrimental impact of using invalid data gradients during training, leading to reduced performance.

\begin{table}[t]
\caption{Comparison of anomaly detection performance: Ours vs the baseline methods (Higher values indicate better performance)}
\label{tab:anomaly_detection_results}
\centering
\begin{tabular}{@{}lccc@{}}
\toprule
Algorithm & Ours & Baseline 1\footnotemark[1] & Baseline 2 \\
\midrule
LOF\footnotemark[2]           & \textbf{83.7} & 78.6 & 79.8 \\
LSAnomaly\footnotemark[3]     & \textbf{91.4} & 86.8 & 87.5 \\
Neural Network                & \textbf{95.9} & 87.1 & 89.6 \\
\botrule
\end{tabular}
\footnotetext[1]{Baseline 1 uses the contractive autoencoder approach~\cite{ml_journal_paper}.}
\footnotetext[2]{LOF: Local Outlier Factor~\cite{lof}.}
\footnotetext[3]{LSAnomaly: Latent Space Anomaly~\cite{lsanomaly}.}
\end{table}

\begin{figure*}[!hbt]
    \centering
    \subfigure[Ours]{
        \includegraphics[trim=0cm 1.5cm 0cm 0cm, clip, width=0.315\textwidth]{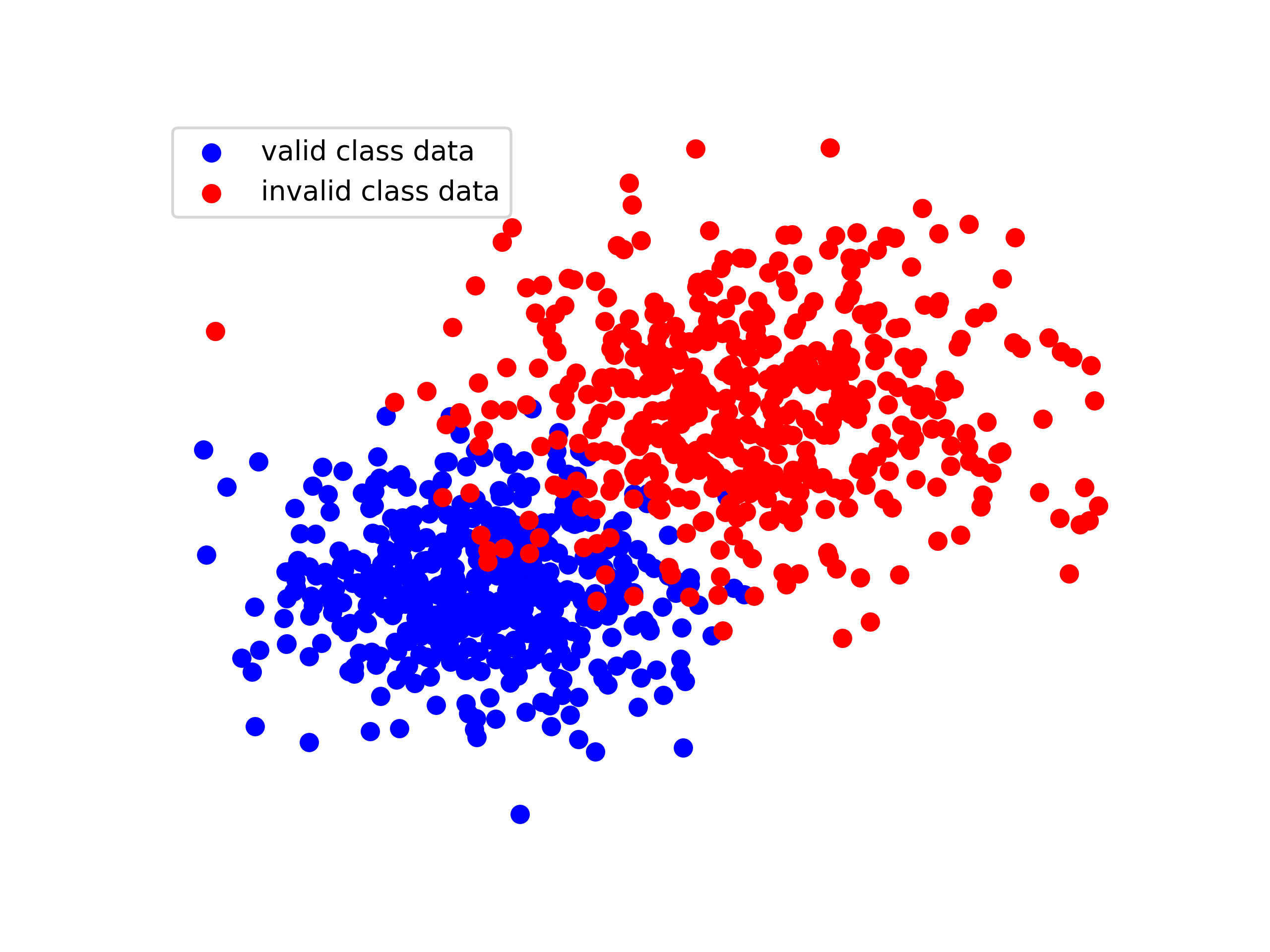}
    }
    \subfigure[Baseline 1 \cite{ml_journal_paper}]{
        \includegraphics[trim=0cm 1.5cm 0cm 0cm, clip, width=0.315\textwidth]{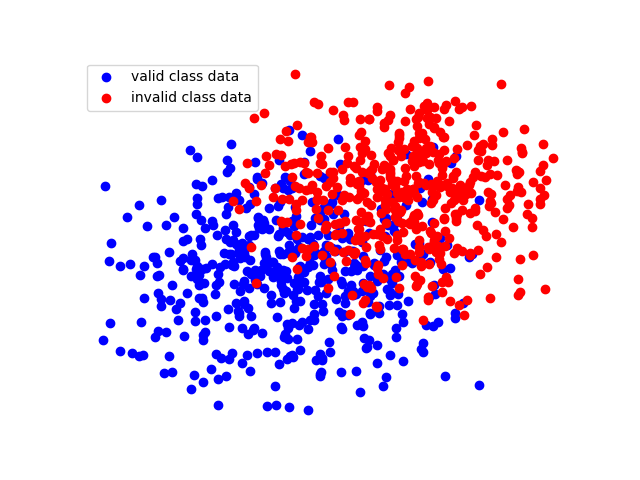}
    }
    \subfigure[Baseline 2]{
        \includegraphics[trim=0cm 1.5cm 0cm 0cm, clip, width=0.315\textwidth]{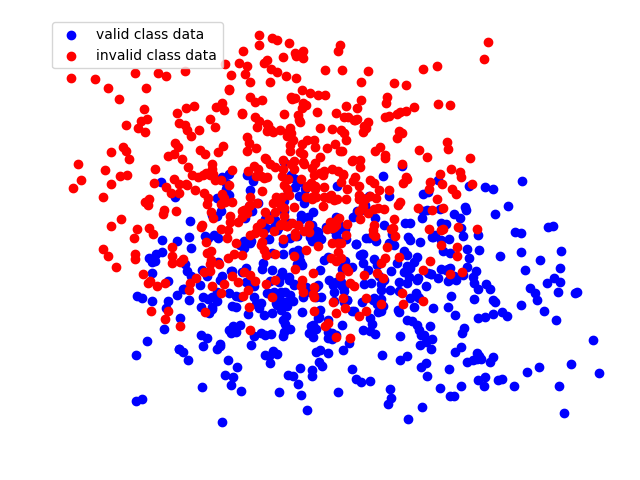}
    }
    \caption{
        t-SNE embeddings of latent representations obtained using (a) the proposed method, (b) Baseline 1 \cite{ml_journal_paper}, and (c) Baseline 2.
    }
    \label{fig:tsne}
\end{figure*}

Furthermore, we visually explore the impact on the clustering and separation of latent representations for valid and invalid classes. Using t-SNE \cite{tsne}, we transform the latent vectors into a two-dimensional space for 500 randomly selected samples of each class in the test dataset. Figure~\ref{fig:tsne} showcases the resulting t-SNE embeddings, with the valid class depicted in blue dots and the invalid class in red. Specifically, Figure~\ref{fig:tsne}(a) illustrates the embeddings acquired using the proposed method while Figures~\ref{fig:tsne}(b) and~\ref{fig:tsne}(c) present the embeddings obtained using the baseline training methods. Upon examination, it is evident that our method leads to a notable improvement in the separation and clustering of the two classes. The clusters in Figure~\ref{fig:tsne}(a) exhibit clearer boundaries and greater distinction compared to Figure~\ref{fig:tsne}(b) and Figure~\ref{fig:tsne}(c).

% \subsubsection{Latent Space Distinctiveness}
We assess the impact of the proposed method on the quality of learned latent representations. We evaluate three metrics: the Bhattacharyya distance between the latent vectors of valid and invalid data samples, the overlap percentage between them, and the distance between their centroids, comparing the proposed method to others.

To compute the overlap percentage, the latent vectors are combined and partitioned into distinct clusters using K-means clustering. 
The number of data points assigned to the incorrect clusters indicates the overlap between the two classes in the latent space. 
The overlap percentage is calculated by dividing the count of overlapping data points by the total number of data points. 
In the centroid distance metric, we determine the centroids for valid and invalid latent vectors by calculating their means. 
The distance between these centroids indicates the separation between the two classes in the latent space. 
A greater distance signifies better distinguishability.

Table \ref{tab:overlap_centroid} compares the Bhattacharyya distance, the overlap percentage and the centroid distance between the proposed method and the baseline methods. For the Bhattacharyya distance metric, our method yields a value of $304.66$, while the baseline methods achieve lower values of $91.88$ and $186.19$. Our method achieves a lower overlap percentage of \SI{11.29}{\percent} compared to \SI{26.41}{\percent} and \SI{18.57}{\percent} for the baseline methods, indicating improved performance. The distance between the centroids is higher for our training method, with a value of $14.90$, compared to the baseline methods, with values $8.57$ and $12.1$.  
The results provide strong evidence that our method leads to higher-quality representations for anomaly detection of DRAM signals.

% \begin{table}[t]
%     \caption{Comparison of Bhattacharyya distance, overlap percentage, and centroid distance: Ours vs. the baseline methods.}
%     \label{tab:overlap_centroid}
%     \centering
%     \begin{tabular}{|l|c|c|c|}
%         \hline
%         \multirow{2}{*}{Metric} & \multicolumn{3}{c|}{Metric Values}
%         \\
%         \cmidrule{2-4}
%         & Ours & Baseline 1 \cite{ml_journal_paper} & Baseline 2\\
%         \hline
%         Bhattacharyya distance ($\uparrow$) & \textbf{304.66} & 91.88 & 186.19 \\
%         Percentage overlap ($\downarrow$)  & \textbf{11.29\%} & 26.41\% & 18.57\% \\
%         Distance between centroids ($\uparrow$) & \textbf{14.90} & 8.57 & 12.1 \\
%         \hline
%     \end{tabular}
% \end{table}

\begin{table}[t]
\caption{Comparison of Bhattacharyya distance, overlap percentage, and centroid distance: Ours vs. the baseline methods.}
\label{tab:overlap_centroid}
\begin{tabular}{@{}llll@{}}
\toprule
Metric & Ours & Baseline 1\footnotemark[1] & Baseline 2 \\
\midrule
Bhattacharyya distance ($\uparrow$) & \textbf{304.66} & 91.88 & 186.19 \\
Percentage overlap ($\downarrow$)   & \textbf{11.29\%} & 26.41\% & 18.57\% \\
Distance between centroids ($\uparrow$) & \textbf{14.90} & 8.57 & 12.1 \\
\botrule
\end{tabular}
\footnotetext[1]{Baseline 1 uses the contractive autoencoder approach~\cite{ml_journal_paper}.}
\end{table}

\subsection{Performance of SI Enhancement Algorithm}
\label{sec:exp:signal_integrity}
The SI enhancement algorithm improves signal integrity by adjusting the signal's latent vector towards an optimal anchor point derived from valid data and then reconstructing the signal with minimal deviation from the original.
To evaluate our SI enhancement algorithm, we use eye diagram analysis. This involves examining eye diagrams for the original and enhanced reconstructed signals. We identify a fixed-height (\SI{80}{\milli\volt}) rectangular window within the eye region and aim to maximize its width while staying within the eye-opening. We quantify the impact of our SI enhancement algorithm by measuring the increase in window area for the enhanced signal compared to the original signal.
To determine the hyperparameter $\alpha$ for our SI enhancement algorithm, we calculate the root mean square error between the input signal and its reconstructed counterpart using fully trained encoder and decoder models with the complete training data. The largest root mean square error observed for the training data is $0.0428$. As a result, we select $\alpha=0.05$.

The results of applying the proposed signal integrity enhancement algorithm to the test dataset are summarized in Table \ref{tab:signal_integrity_improvement_proposed_baseline_comparison}. 
The table provides statistics on the percentage improvement in window areas within the eye diagram, which serves as a measure of signal integrity enhancement. Our algorithm achieves an \SI{11.30}{\percent} average improvement in the window area, indicating its efficacy in enhancing signal integrity. With a low standard deviation of 3.41, the majority of signals experience substantial improvements, leading to more reliable and accurate signal enhancement. The maximum improvement observed is \SI{31.8}{\percent}, showing the potential of our algorithm to greatly enhance signal integrity in specific scenarios. The minimum improvement of \SI{0.88}{\percent} demonstrates a consistent positive impact on signal integrity. The median improvement of \SI{7.71}{\percent} further supports the effectiveness of the algorithm, as a significant proportion of signals achieved improvements above this value.

\begin{figure*}[htbp]
    \centering
    \subfigure[Original signal eye diagram]{
        \includegraphics[width=0.45\textwidth]{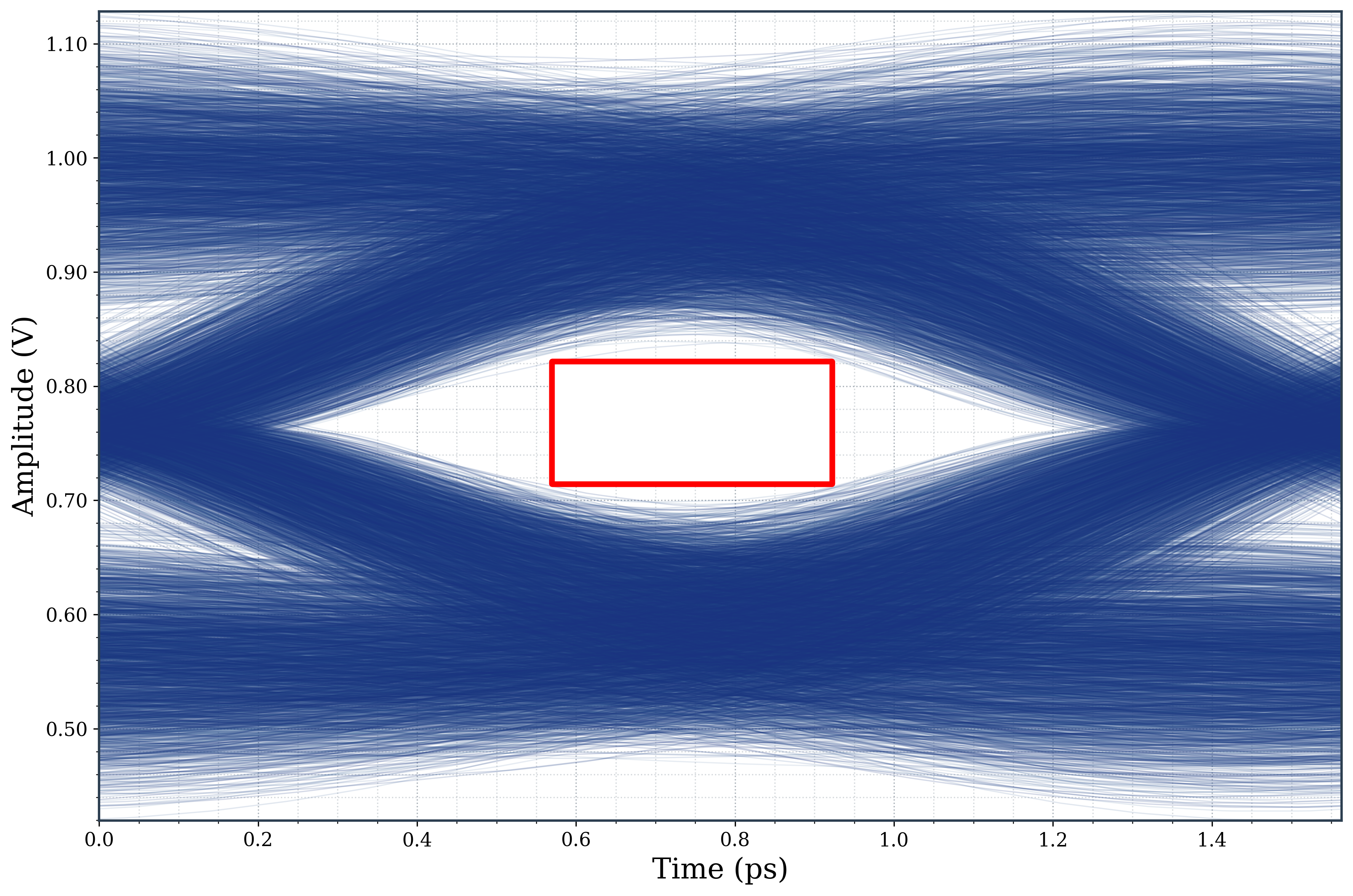}
    }
    \subfigure[Eye Diagram after the application of SI enhancement algorithm]{
        \includegraphics[width=0.45\textwidth]{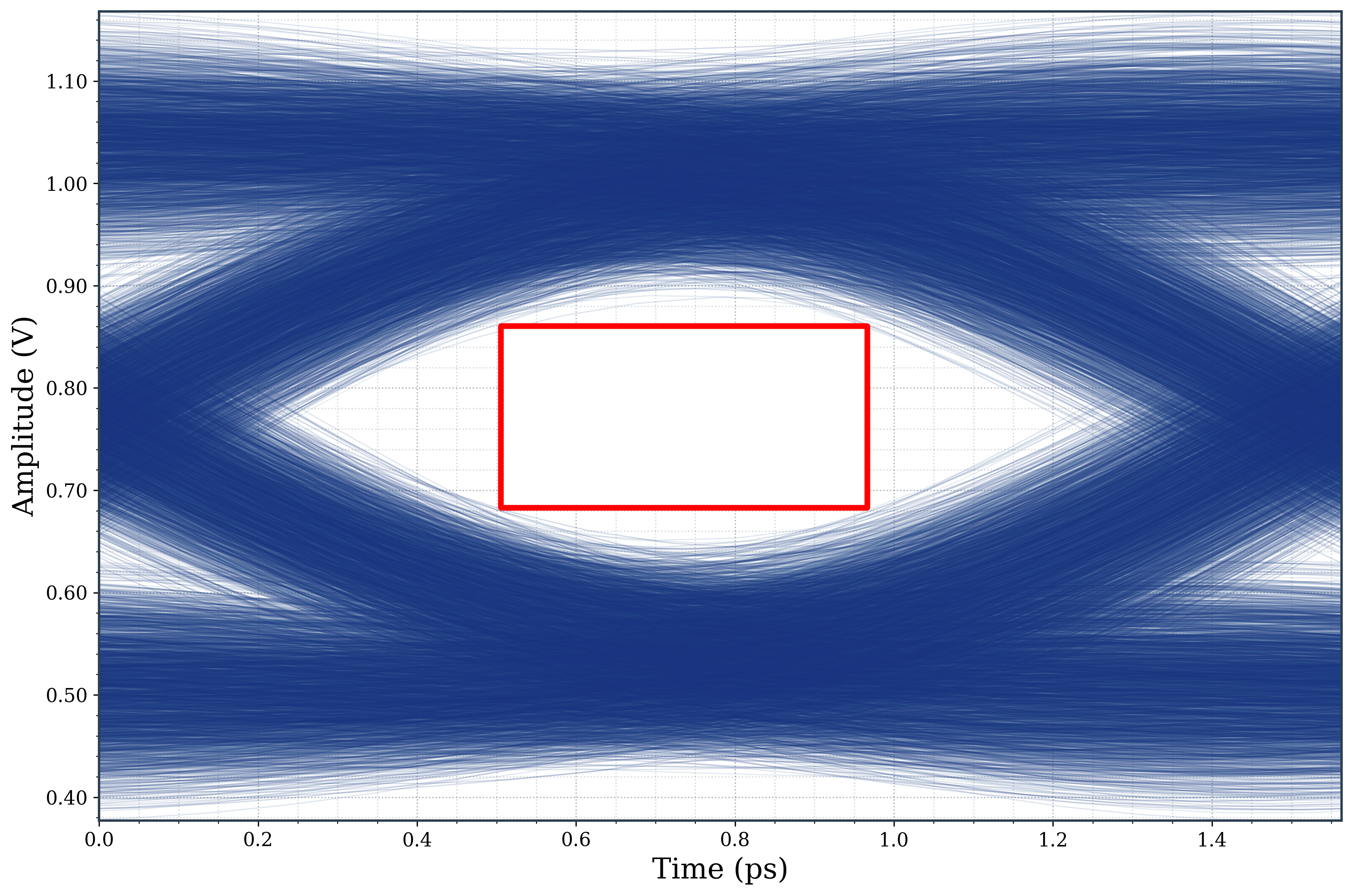}
    }
    \caption{Comparison of eye diagrams before and after the application of our proposed SI enhancement algorithm. For this example, the largest rectangular window area within the eye-opening (shown in red) increases from \SI{39.1}{\milli\volt\pico\second} (original signal) to \SI{59.4}{\milli\volt\pico\second} (SI enhanced signal), corresponding to a \SI{51.9} increase in area.}
    \label{fig:eyediagrams}
\end{figure*}

The effectiveness of the proposed SI enhancement algorithm is highly dependent on the quality of the latent vectors it utilizes. Consequently, we evaluate and compare the performance of latent representations derived from the proposed training method against those obtained from baseline training methods, specifically focusing on signal integrity improvement. Table \ref{tab:signal_integrity_improvement_proposed_baseline_comparison} shows statistics on the percentage improvement in window area for both methods on the test dataset. Our method achieves a higher mean percentage improvement of \SI{11.3}{\percent}, with a lower standard deviation of \SI{3.41}{\percent}, compared to the baseline methods. Our method also achieves a higher maximum improvement of \SI{31.8}{\percent} and a higher minimum improvement of \SI{0.88}{\percent} compared to the baseline methods. The median improvement for our method is \SI{7.71}{\percent},  exceeding that of both baseline methods. These results demonstrate the superior performance of latent representations obtained using the proposed training method in enhancing signal integrity compared to the baseline methods.

% \begin{table}[t]
%     \caption{Comparison between the proposed and the baseline training methods for SI enhancement: percentage improvement in the window area.}
%     \label{tab:signal_integrity_improvement_proposed_baseline_comparison}
%     \centering
%     \begin{tabular}{|l|c|c|c|}
%     \hline
%     \multirow{2}{*}{Metric} & \multicolumn{3}{c|}{Percentage improvement in window area (\%)}
%     \\
%     \cmidrule{2-4}
%      & Ours & Baseline 1 \cite{ml_journal_paper} & Baseline 2\\
%      \hline
%      % \hline
%       Mean $(\uparrow)$  & \textbf{11.30} & 6.35 & 8.94 \\
%       Std $(\downarrow)$ & \textbf{3.41} & 4.88 & 5.11\\
%       Maximum $(\uparrow)$ & \textbf{31.8} & 25.1 &  28.3\\
%       Minimum $(\uparrow)$ & \textbf{0.88} &  0.33 & 0.61\\
%       Median $(\uparrow)$ & \textbf{7.71} & 5.76 & 7.06\\
      
%    \hline
%     \end{tabular}
% \end{table}

\begin{table}[t]
\caption{Comparison between the proposed and the baseline training methods for SI enhancement: percentage improvement in the window area.}
\label{tab:signal_integrity_improvement_proposed_baseline_comparison}
\centering
\begin{tabular}{@{}lccc@{}}
\toprule
\multirow{2}{*}{Metric} & \multicolumn{3}{c}{Percentage improvement in window area (\%)} \\
\cmidrule{2-4}
 & Ours & Baseline 1\footnotemark[1] & Baseline 2 \\
\midrule
Mean $(\uparrow)$    & \textbf{11.30} & 6.35 & 8.94 \\
Std $(\downarrow)$   & \textbf{3.41}  & 4.88 & 5.11 \\
Maximum $(\uparrow)$ & \textbf{31.8}  & 25.1 & 28.3 \\
Minimum $(\uparrow)$ & \textbf{0.88}  & 0.33 & 0.61 \\
Median $(\uparrow)$  & \textbf{7.71}  & 5.76 & 7.06 \\
\botrule
\end{tabular}
\footnotetext[1]{Baseline 1 uses the contractive autoencoder approach~\cite{ml_journal_paper}.}
\end{table}

Figure~\ref{fig:eyediagrams} presents a qualitative comparison of eye diagrams before and after applying our SI enhancement algorithm to an example signal. After applying our SI enhancement algorithm, we observe a significant improvement in signal quality, as shown in Figure~\ref{fig:eyediagrams}(b). For this particular example, the rectangular window area within the eye-opening, showon in red, increased by \SI{51.9}{\percent}. Specifically, the largest rectangular window area in the original signal is \SI{39.1}{\milli\volt\pico\second}, which increases to \SI{59.4}{\milli\volt\pico\second} after SI enhancement.

\subsection{Ablation Studies}

\begin{figure}[t]
    \centering
    \includegraphics[width=.75\linewidth]{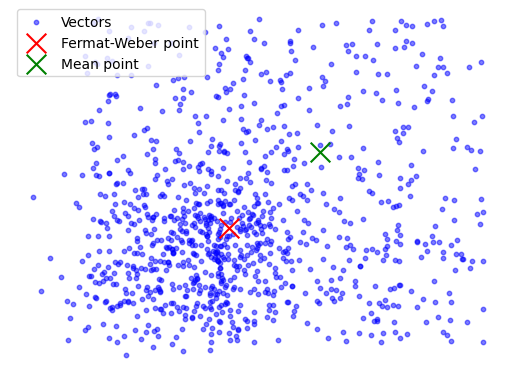}
    \caption{t-SNE embeddings of 1000 latent vectors from the valid class (depicted in blue), accompanied by the Fermat-Weber point in red and the mean point in green.}
    \label{fig:fermat}
\end{figure}

% \begin{table}[ht]
%     \caption{Comparison of anomaly detection performance with latent vectors trained on valid vs. invalid data gradients}
%     \label{tab:valid_invalid_combined}
%     \centering
%     \begin{tabular}{|l|c|c|c|c|}
%     \hline
%     \multirow{2}{*}{Algorithm} & \multicolumn{2}{c|}{Accuracy (\%)} & \multicolumn{2}{c|}{True Negative Rate (\%)} \\
%     \cmidrule{2-5}
%      & \makecell[c]{Valid\\gradients} & \makecell[c]{Invalid\\gradients} & \makecell[c]{Valid\\gradients} & \makecell[c]{Invalid\\gradients} \\
%      \hline
%       LOF \cite{lof}             & \textbf{81.4} & 57.3 & \textbf{78.3} & 52.6 \\
%       LSAnomaly \cite{lsanomaly} & \textbf{85.7} & 61.0 & \textbf{81.3} & 52.7 \\
%       Neural Network             & \textbf{93.1} & 76.3 & \textbf{88.7} & 63.5 \\
%       \hline
%     \end{tabular}
% \end{table}

\begin{table}[h]
\caption{Comparison of anomaly detection performance with latent vectors trained on valid vs. invalid data gradients}
\label{tab:valid_invalid_combined}
\centering
\begin{tabular}{@{}lcccc@{}}
\toprule
\multirow{2}{*}{Algorithm} & \multicolumn{2}{c}{Accuracy (\%)} & \multicolumn{2}{c}{True Negative Rate (\%)} \\
\cmidrule(lr){2-3} \cmidrule(lr){4-5}
 & Valid gradients & Invalid gradients & Valid gradients & Invalid gradients \\
\midrule
LOF\footnotemark[1]             & \textbf{81.4} & 57.3 & \textbf{78.3} & 52.6 \\
LSAnomaly\footnotemark[2]       & \textbf{85.7} & 61.0 & \textbf{81.3} & 52.7 \\
Neural Network                  & \textbf{93.1} & 76.3 & \textbf{88.7} & 63.5 \\
\botrule
\end{tabular}
\footnotetext[1]{LOF: Local Outlier Factor~\cite{lof}.}
\footnotetext[2]{LSAnomaly: Latent Space Anomaly~\cite{lsanomaly}.}
\end{table}

\subsubsection{Choice of the Anchor Point}
We opt for the Fermat-Weber point as the anchor, in contrast to using the mean of the latent vectors belonging to the valid class.
This choice is motivated by the ability of the Fermat-Weber point to handle outliers effectively. 
In Figure~\ref{fig:fermat}, the t-SNE embeddings of 1000 randomly sampled valid class latent vectors from the test dataset are depicted in blue, accompanied by the t-SNE embeddings of the Fermat-Weber point in red, and the mean point in green. Notably, the red point is positioned closer to the cluster of latent vectors, demonstrating its effectiveness in mitigating the impact of outliers.

\subsubsection{Relationship Between SI Metric Values and Signal Integrity}
\label{sec:exp:si_metric}
\begin{figure}[t]
    \centering
    \includegraphics[width=.75\textwidth]{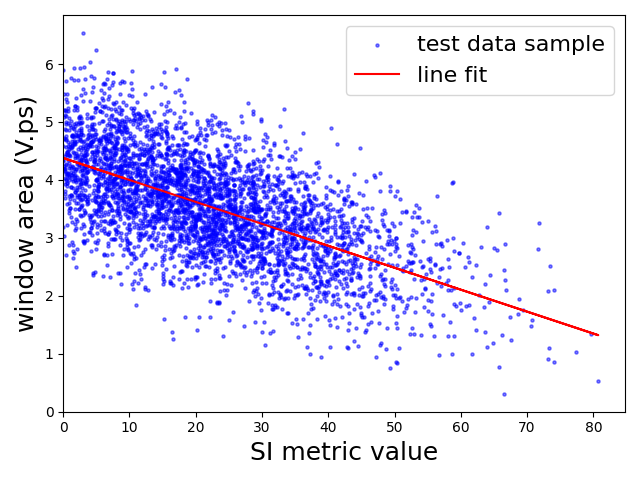}
    \caption{Relationship between SI metric values and window area}
    \label{fig:window_area}
\end{figure}

We analyze the relationship between the SI metric value and signal integrity to demonstrate that our SI metric accurately reflects signal quality.
In Figure~\ref{fig:window_area}, the SI metric values are plotted against the corresponding window area for the entire test dataset. Additionally, a line fit is included in the plot to demonstrate the trend. The line equation, $w = -0.0379s + 4.3783$ with $w$ representing the window area of a signal with SI metric value $s$, exhibits a negative slope. The negative slope of $-0.0379$ clearly demonstrates that as the SI metric value increases, the window area decreases. These observations confirm that lower values of the SI metric correspond to higher signal integrity.

\subsubsection{Impact of Training with Valid vs. Invalid Data Gradients on Anomaly Detection Performance}

We investigate whether using only valid data gradients versus only invalid data gradients during training leads to better performance.
For this experiment, we exclusively utilize newly partitioned training and test datasets based on waveform generation conditions. 
The training data comprises ISI-affected data corresponding to cases 1-4 of the five different conditions, while the test data consists of crosstalk-affected data from case 5.
Two models combining an autoencoder and a classifier are trained: one with only valid data gradients and the other with only invalid data gradients. 
We then evaluate the quality of latent vectors extracted by the trained encoders on the test set using the three anomaly detection methods.
The loss function for training using valid data gradients is given in equation \eqref{eq:loss} while the loss function for training with invalid data gradients is given as 
$\mathcal{L} =  \Vert \mathbf{x} - \hat{\mathbf{x}} \Vert^{2} - (1 - y) \log(1 - \hat{y})$.

Table \ref{tab:valid_invalid_combined} compares the performance of different anomaly detection algorithms, comparing accuracy and true negative rate which represents instances where the anomaly detection algorithm accurately identifies data samples as belonging to the invalid class. 
Results from Table \ref{tab:valid_invalid_combined} indicate that latent vectors obtained using valid data gradients yield superior performance compared to those obtained using invalid data gradients across all anomaly detection algorithms. This is also evident in true negative rates, where latent vectors obtained using valid data gradients consistently outperform those obtained using invalid data gradients.

One potential explanation for this difference in performance could be the challenge of comprehensively gathering training data covering all possible anomalies. However, we can typically collect data that aligns with the characteristics of valid data samples. Training with invalid data gradients, which predominantly correspond to a specific class of anomaly, may lead to overfitting. Consequently, this could result in poorer performance when detecting anomalies of other types. 
Therefore, the proposed factor that only uses valid data gradients enables the generation of high-quality latent vectors for anomaly detection.
\section{Conclusions}
\label{sec:conclusions}

We present a framework integrating an autoencoder and classifier that significantly enhances anomaly detection and signal integrity improvement in high-speed DRAM signals. By focusing on valid data features, our approach generates more distinctive latent representations, improving anomaly detection accuracy to 95.9\% with a neural network detector, a 5-9\% gain over baseline methods. The learned latent space exhibits superior class separation, evidenced by a Bhattacharyya distance of 304.66 (vs. $<$187 for baselines) and a reduced overlap of 11.29\% (vs. $>$18\%) in the latent space. Additionally, our signal integrity enhancement algorithm achieves an average 11.3\% improvement in eye diagram window area (upto 31.8\% for some signals), mitigating signal distortions.

\bibliography{references}% common bib file

\backmatter
\section*{Statements and Declarations}
\bmhead{Acknowledgements}
This work was supported by Samsung Electronics Co. Ltd (Contract ID : MEM230315\_0004).

\bmhead{Author Contributions}
Muhammad Usama conceptualized the study, developed the proposed methods, and conducted the experiments. Hee-Deok Jang contributed to the conceptualization, manuscript writing and data processing. Soham Shanbhag and Dong Eui Chang assisted in refining the manuscript and enhancing the experimental evaluation. Yoo-Chang Sung and Seung-Jun Bae were responsible for data collection. All authors reviewed and approved the final manuscript.

\end{document}